\pdfoutput=1

\documentclass[11pt]{article}

\usepackage{EMNLP2022}

\usepackage{times}
\usepackage{latexsym}
\usepackage{booktabs,times}
\usepackage{latexsym}
\usepackage{amssymb}
\usepackage{xspace}
\usepackage{bibentry}
\usepackage{amssymb}
\usepackage{subcaption}
\usepackage{lineno}
\usepackage{algorithm}
\usepackage{algorithmic}
\usepackage{enumerate}
\usepackage{amsmath}
\usepackage{graphicx}
\usepackage{multirow}
\usepackage{url}
\usepackage{tabu}
\usepackage{bm}
\usepackage{dirtytalk}
\usepackage{multirow}
\usepackage{tabularx}
\usepackage{array}
\usepackage{cite}
\usepackage{booktabs}
\usepackage{color}
\usepackage{wasysym}
\usepackage{pifont}
\usepackage{hyperref}
\usepackage{booktabs}
\usepackage{CJKutf8}
\usepackage{pifont}%
\newcommand{\xmark}{\ding{55}}%
\usepackage{euflag}

\usepackage{CJKutf8}

\usepackage[T5]{fontenc}

\usepackage[utf8]{inputenc}

\usepackage{microtype}

\usepackage{inconsolata}

\newcommand{\vl}{V\&L}
\newcommand\doublecheck{{\checked\kern-0.5em\checked}}

\newcommand{\invisible}[1]{}

\title{Multilingual Multimodal Learning with Machine Translated Text}

\usepackage{tipa}

\newcommand{\wust}{\text{\normalfont \textipa{A}}}
\newcommand{\upb}{\text{\normalfont \textipa{B}}}
\newcommand{\ku}{\text{\normalfont \textipa{C}}}
\newcommand{\pcai}{\text{\normalfont \textipa{D}}}

\author{
  Chen Qiu$^{\wust}$ \ \ 
  Dan Oneață$^{\upb}$ \ \ 
  Emanuele Bugliarello$^{\ku}$ \\
  {\bf%
  Stella Frank$^{\ku,\pcai}$ \ \
  Desmond Elliott$^{\ku,\pcai}$} \\
  $^{\wust}$School of Computer Science and Technology,\\ Wuhan University of Science and Technology, China \\
  $^{\upb}$ University Politehnica of Bucharest, Romania \\
  $^{\ku}$Department of Computer Science, University of Copenhagen, Denmark \\
  $^{\pcai}$Pioneer Centre for AI, Denmark \\
  \texttt{chen@wust.edu.cn \  dan.oneata@speed.pub.ro \ \{emanuele, stfr, de\}@di.ku.dk} 
}

\begin{document}
\maketitle

\begin{abstract}
Most vision-and-language pretraining research focuses on English tasks. However, the creation of multilingual multimodal evaluation datasets (e.g. Multi30K, xGQA, XVNLI, and MaRVL) poses a new challenge in finding high-quality training data that is both multilingual and multimodal.
In this paper, we investigate whether machine translating English multimodal data can be an effective proxy for the lack of readily available multilingual data.
We call this framework TD-MML: Translated Data for Multilingual Multimodal Learning, and it can be applied to any multimodal dataset and model. We apply it to both pretraining and fine-tuning data with a state-of-the-art model. In order to prevent models from learning from low-quality translated text, we propose two metrics for automatically removing such translations from the resulting datasets.
In experiments on five tasks across 20 languages in the IGLUE benchmark, we show that translated data can provide a useful signal for multilingual multimodal learning, both at pretraining and fine-tuning.

\end{abstract}

\section{Introduction}
\label{sec:introduction}

Vision-and-language (\vl) pretraining is the process of learning deep contextualised cross-modal representations from large collections of image--sentence pairs~\citep[\textit{inter-alia}]{li2019visualbert,tan-bansal-2019-lxmert,chen2020eccv}. 
These pretrained models are an excellent backbone for transfer learning to a wide range of downstream tasks, such as visual question answering~\citep{antol2015vqa,Gurari_2018_CVPR,agrawal2022rethinking}, referring expression alignment~\citep{kazemzadeh-etal-2014-referitgame,7780378}, and image--sentence retrieval~\citep{young2014image,Lin2014eccv}. 
Thus far, downstream evaluations have mostly focused on English tasks due to the availability of datasets, but the recent IGLUE benchmark~\citep{bugliarello2022} makes it now possible to evaluate models on several downstream tasks across 20 languages. 

\begin{figure}
    \centering
    \includegraphics[width=0.45\textwidth, trim={4cm 6.5cm 10.5cm 1.5cm}, clip]{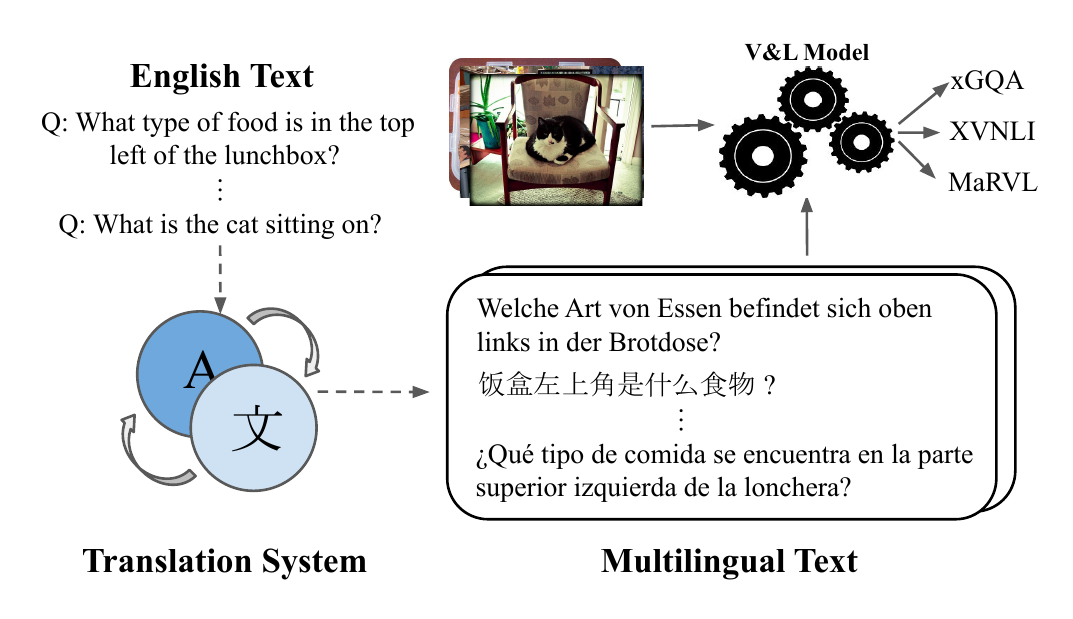}
    \vspace{-3mm}
    \caption{Multilingual multimodal data is a scarce resource compared to English multimodal data. Given an English multimodal dataset, we generate a multilingual dataset using a black box translation system. We explore the utility of this approach to creating multilingual text for both downstream task fine-tuning and pretraining.}\label{fig:intro:overview}
    \vspace{-4mm}
\end{figure}

Success in multilingual multimodal tasks, such as those in IGLUE, is expected to depend on models with grounded representations that transfer across languages~\citep{bugliarello2022}. 
For example, in the MaRVL dataset~\citep{liu-etal-2021-visually}, models need to deal with a linguistic and cultural domain shift compared to English data.
Therefore, an open problem is to define pretraining strategies that induce high-quality multilingual multimodal representations. 
Existing work has tackled this problem by either jointly training on English multimodal data and multilingual text-only data~\citep{liu-etal-2021-visually,Ni_2021_CVPR}, pretraining with a private dataset of multilingual captioned images~\citep{jain-etal-2021-mural-multimodal}, or machine translating multimodal pretraining data~\citep{Zhou_2021_CVPR}. 

In this paper, we further investigate the potential of machine translated text for both fine-tuning \emph{and} pretraining across four diverse \vl{} tasks.\footnote{The models and the machine translated text are available at 
\url{https://github.com/danoneata/td-mml}}
The overarching motivation is that machine translation is an inexpensive approach to producing large amounts of multilingual text compared to collecting data from humans, or scraping high-quality image--caption pairs from the web. 
Having access to thousands of data points in a target language might indeed be necessary to improve cross-lingual performance in downstream tasks~\citep{bugliarello2022}.
As such, translating fine-tuning data into multiple languages may be a compelling approach towards downstream task success.
Moreover, if this can be achieved through machine translated text, it raises the question of whether we can also pretrain on many millions of multilingual translated examples. 
Motivated by the initial experiments of \citet{Zhou_2021_CVPR}, we test this hypothesis further, on more languages and more tasks, reporting more nuanced results from large-scale translated text.

Overall, we show that machine translation can provide inexpensive and impressive improvements when fine-tuning models for multilingual multimodal tasks. Moreover, translation-based pretraining leads to significant gains in zero-shot cross-lingual transfer over existing approaches. However, we find mixed results when combining this with multilingual fine-tuning. There are still opportunities to realise further benefits from machine translated text, which may be found through more compute-intensive pretraining.

\textbf{Contributions}. \textbf{1)} We present the TD-MML framework to narrow the gap between English and non-English languages in multimodal research. \textbf{2)} In the process of translation-based pretraining, we present a reliable strategy to filter out bad translations.  \textbf{3)}  We conduct systematic evaluations in zero-shot and machine translated scenarios, and show the benefits that can be gained from simply having more data in the target languages.

\section{Related Work}
\label{sec:related-work}

Inspired by the success of self-supervised language model 
pretraining~\citep[\textit{inter-alia}]{devlin-etal-2019-bert}, researchers have also explored this paradigm 
with multimodal models~\citep{gella-etal-2017-image, akoskadar-conll-2018-image}. The first 
wave~\citep{li2019visualbert,tan-bansal-2019-lxmert,Li2020oscar,chen2020eccv} were initialised from BERT and 
pretrained on English image--text datasets like Conceptual Captions~\citep{sharma-etal-2018-conceptual} and 
COCO~\citep{Lin2014eccv}, where the visual modality was represented using feature vectors extracted from 
10--100 automatically detected object proposals~\citep{8578734}. %
More recent models~\citep{kim2021vilt,li-etal-2021-alignbefore,Singh_2022_CVPR} represent the visual modality
using a Vision Transformer~\citep{dosovitskiy-etal-animage}, which can be end-to-end fine-tuned during 
pretraining, as opposed to working with pre-extracted object proposals.

More related to our work are the multilingual variants of these 
models~\citep{liu-etal-2021-visually,Zhou_2021_CVPR,Ni_2021_CVPR,jain-etal-2021-mural-multimodal}. 
The lack of large-scale multilingual multimodal datasets has resulted in different strategies to train such 
models. 
\citet{liu-etal-2021-visually} simply augment English caption data with text-only multilingual Wikipedia 
data.
In addition to this, \citet{Ni_2021_CVPR} further create code-switched multimodal data\footnote{French 
code-switching might transform ``The dog is  chasing a ball'' into ``The \emph{chien} \emph{est} chasing a 
ball'', for example.} by randomly swapping English words in Conceptual Captions with the corresponding 
translation in one of 50 other languages, obtained through the \textit{Panlex} dictionary.
On the other hand, \citet{Zhou_2021_CVPR} machine translate the Conceptual Captions dataset into German, 
French, Czech, Japanese, and Chinese, for a total of 19.8M pretraining data points. 
Finally, \citet{jain-etal-2021-mural-multimodal} pretrain on 3.6B multilingual captions by extending the 
Conceptual Captions collection pipeline to multiple languages.\footnote{This large-scale dataset is not 
publicly available.}

In this paper, we further explore the potential of machine translation for pretraining and fine-tuning.
\citet{Zhou_2021_CVPR} first pretrained a model on machine translations of the Conceptual Captions pretraining data in five 
high-resource languages (Mandarin Chinese, Czech, French, German, and Japanese), which then resulted in 
overall better multilingual representations across a number of diverse languages~\citep{bugliarello2022}.
Here, we explore the potential of training multimodal models on a much larger and diverse set of languages, 
including low-resource ones. Effectively doing so requires tackling issues and limitations with machine 
translation systems, which do not produce high quality translations across all languages. This is especially 
relevant when translating a large corpus, which might include a large number of data 
points with low-quality text.

\section{The IGLUE Benchmark}

The impetus of our work is the recent creation of the Image-Grounded Language Understanding Evaluation (IGLUE; \citealt{bugliarello2022}) benchmark for evaluating multimodal models across twenty languages and four tasks, using five different datasets.
Specifically, the benchmark focuses on zero- and few-shot transfer, where models are fine-tuned on English data and then tested to cross-lingually generalise with no or few samples in the target language for the target downstream task. 
The following datasets are included in IGLUE:

\begin{description}
    \item[XVNLI]\hspace{-1ex} is a cross-lingual Visual Natural Language Inference task~\citep{bugliarello2022}, which requires models to predict the relation (entailment, contradiction, or neutral) between a premise in the form of an image, and a hypothesis in the form of a sentence.
    \item[xGQA]\hspace{-1ex} is a cross-lingual Grounded Question Answering task~\citep{pfeiffer2021xgqa},
    using images from Visual Genome \citep{10.1007/s11263-016-0981-7} and translations of the English questions from GQA \citep{Hudson_2019_CVPR}. The questions in GQA are automatically generated from the image scene graphs. %
    \item [MaRVL]\hspace{-1ex} focuses on multicultural reasoning over images~\citep{liu-etal-2021-visually}. The task is in the same format as the English NLVR2 \citep{suhr-etal-2019-corpus} data, namely to judge whether a sentence is true or false for a pair of images. However, the images and the descriptions are sourced directly in the target languages. %
    \item[xFlickr\&CO]\hspace{-1ex} is an evaluation dataset for image--text retrieval in eight high-resource languages~\citep{bugliarello2022}. The images are collected from Flickr30K \citep{young2014image} and COCO \citep{Lin2014eccv}, while the captions are new descriptions sourced from native speakers in the target languages.
    \item[WIT] is a second image--text retrieval dataset based on the Wikipedia-based Image Text dataset~\citep{srinivasan2021-SIGIR}. WIT is scraped directly from Wikipedia and contains a much more diverse set of image types than the other datasets, as well as more complex and entity-centric descriptions.
\end{description}

Each of the tasks has a natural English training counterpart: SNLI-VE \citep{Xie2019NLIVE} for XVNLI; GQA for xGQA, NLVR2 for MaRVL, and English training splits of Flickr30K and WIT.

\citet{bugliarello2022} found that current multilingual \vl{} models show a large gap in performance, in each of these tasks, when evaluating on non-English data.
Moreover, further training these models on a few examples in a target language only slightly improved their cross-lingual capabilities.

\section{Fine-Tuning with Translated Data}
\label{sec:xuniter-finetune}

As an initial experiment, we investigate the extent to which performance can be improved by fine-tuning on multilingual machine-translated data instead of only English data. We conduct this experiment on the MaRVL and xGQA datasets. The results can be seen in Tables~\ref{xuniter-marvl} and \ref{xuniter-xgqa}, respectively. %

\begin{table}[t]
	\centering
	\resizebox{\linewidth}{!}{
    \begin{tabular}{lr|rrrrrrr}
		\toprule
		\textbf{Approach} & \textbf{ENG} & \textbf{IND} & \textbf{SWA} & \textbf{TAM} & \textbf{TUR} & \textbf{CMN}  & \textbf{avg}\\
		\midrule
		English & 71.6 & 55.1 & 55.5 & 53.1 & 56.2 & 53.1 & 54.6 \\  %
		MT      & 67.9 & 59.6 & 61.4 & 60.4 & 64.3 & 59.4 & 61.0  \\  %
		\bottomrule
	\end{tabular}
	}
	\caption{
    MaRVL accuracy results for zero-shot cross-lingual evaluation, i.e. English-only NLVR2 fine-tuning, and multilingual fine-tuning using machine translated NLVR2 data (MT).
    The average results exclude ENG accuracy.
    }
    \label{xuniter-marvl}
\end{table}

\begin{table}[t]
	\centering
	\resizebox{\linewidth}{!}{
    \begin{tabular}{lr|rrrrrrrr}
		\toprule
		\textbf{Approach} & \textbf{ENG} & \textbf{BEN} & \textbf{DEU} & \textbf{IND} & \textbf{KOR} & \textbf{POR} & \textbf{RUS}  & \textbf{CMN} & \textbf{avg}\\
		\midrule
		English     & 54.8 & 10.8 & 34.8 & 33.7 & 12.1  & 22.1  & 18.8 & 19.6 & 21.7\\  %
		MT      & 48.1 & 41.8 & 46.5 & 45.7 & 44.8 & 46.8 & 46.2 & 45.7 & 45.3  \\  %
		\bottomrule
	\end{tabular}
	}
	\caption{
    xGQA accuracy results for zero-shot cross-lingual evaluation, i.e. English-only GQA fine-tuning, and multilingual finetuning using machine translated GQA data (MT).
    Average results exclude ENG accuracy.%
    }
    \label{xuniter-xgqa}
\end{table}

We use the M2M-100-large model \citep{fan2021jmlr} 
to translate the NLVR2 training data into the 5 MaRVL languages,
and the GQA training data into the 7 xGQA languages.
For the model, we use the xUNITER 
\citep{liu-etal-2021-visually} implementation from VOLTA%
~\citep{bugliarello-etal-2021-multimodal}.
xUNITER extends the UNITER architecture~\citep{chen2020eccv} multilingually, by initializing the model from XLM-RoBERTa~\citep{conneau-etal-2020-unsupervised} and pretraining on English image captions and text-only multilingual Wikipedia paragraphs. 
Starting from the publicly-released xUNITER checkpoint, we fine-tune on the machine translated training sets for each task.
For a fair comparison to English-only fine-tuning, %
we ensure that the multilingual fine-tuning is based on the same number of parameter updates. In effect, this reduces the number of training epochs from 20$\rightarrow$3 for MaRVL, and 5$\rightarrow$1 for xGQA. We round number of epochs so it is close to the English-only setup.\footnote{Note: This is an approximation. For MaRVL, 3 epochs are equivalent to 18 (rather than 20) of English-only data (6 languages). For xGQA, 1 translated text epoch is equivalent to 8 English-only epochs (8 languages) rather than 5.} %
This means that, in our setup, all the images are seen for the same number of times, but each unique caption will be seen fewer times in each of the target languages.

Using machine translated data for fine-tuning brings large improvements in performance for both MarVL and xGQA. Table~\ref{xuniter-marvl} shows the results for MaRVL, where each non-English language increases by between 4.5--8.1 accuracy points. Table~\ref{xuniter-xgqa} shows the results for xGQA, where the performance for the non-English languages increases by 11.7--32.7 points.
We also observe small decreases in performance on English for each task but this is expected. Recall that the models were fine-tuned for the same number of steps, which means the model fine-tuned on translations has been exposed to less English text in order to process multilingual text.
We conclude that using machine translated fine-tuning data is an inexpensive and viable path to better task-specific performance.

\section{On Pretraining with Translated Data}

The previous section showed the benefits of using machine translated data for multilingual fine-tuning. We now turn our attention to whether further improvements can be realised by adding multilinguality via machine translated data is useful for pretraining.
This requires two components: (i) a large-scale translation pipeline and the means to deal with potential data quality issues, and (ii) a model that can exploit the machine translated training data, which we dub TD-MML for Translated-Data Multimodal Multilingual Learning.

\subsection{Translation and Data Preparation}
\label{subsec:translation}

\begin{figure}[t]
    \begin{subfigure}[t]{0.48\textwidth}
        \includegraphics[width=\textwidth]{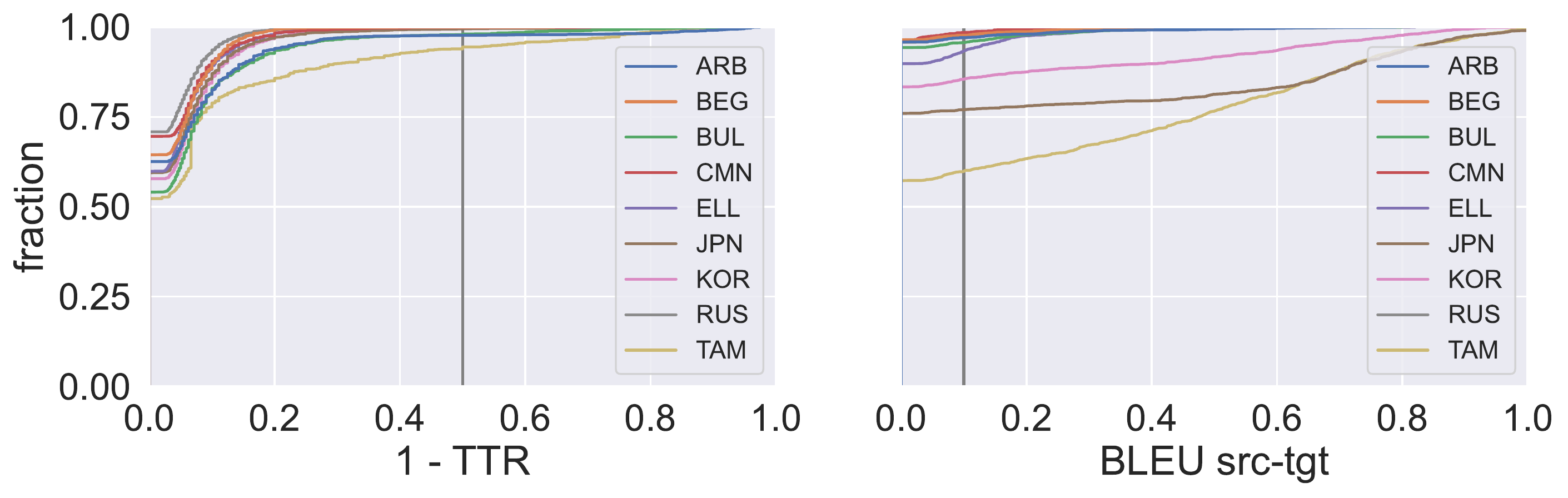}
        \subcaption{languages with a non-Latin script}
    \end{subfigure}
    \begin{subfigure}[t]{0.48\textwidth}
        \includegraphics[width=\textwidth]{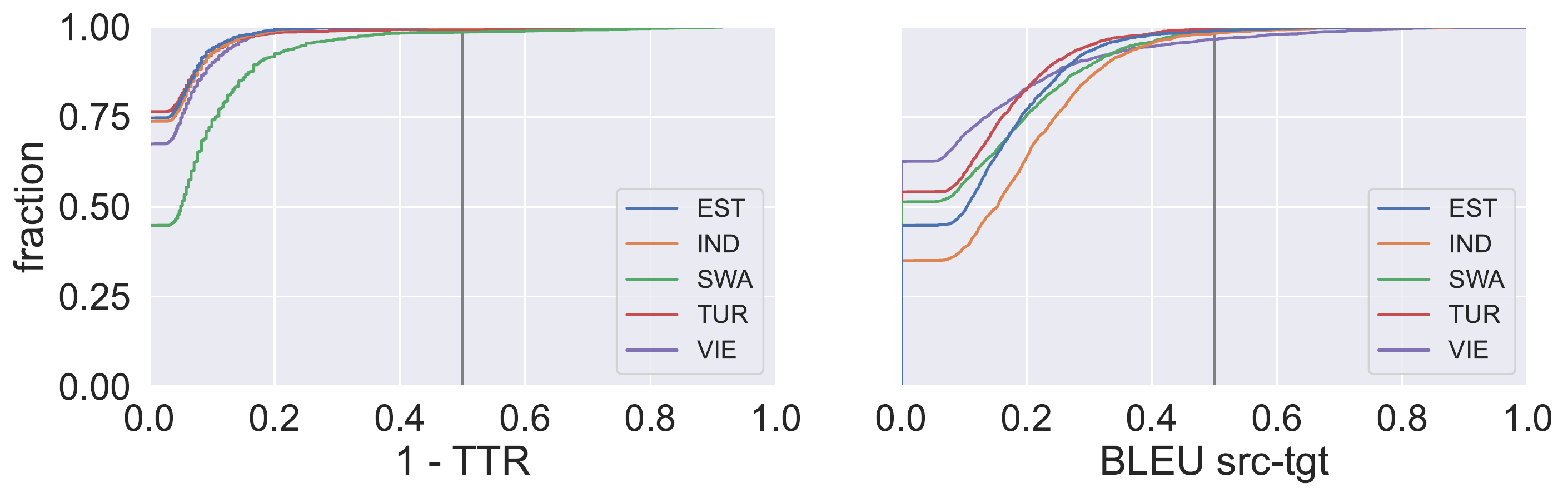}
        \subcaption{non-Indo-European languages}
    \end{subfigure} 
    \begin{subfigure}[t]{0.48\textwidth}
        \includegraphics[width=\textwidth]{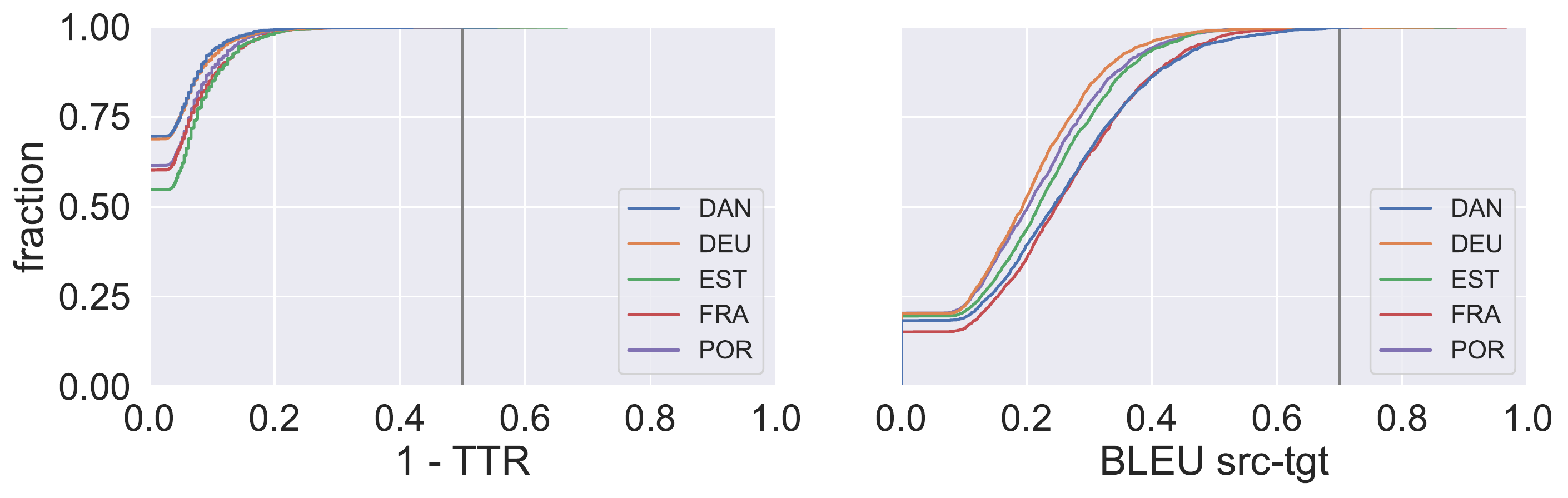}
        \subcaption{Indo-European languages with Latin script}
    \end{subfigure}
    \caption{%
    Cumulative distributions of the two badness scores (1 - TTR, the complement of the token-to-type ratio, and BLUE src-tgt, the BLEU score between the source and target sentence) for the nineteen non-English languages in IGLUE.
    The languages are grouped in three categories, and the vertical lines denote the filtering thresholds for each of the categories and two scores.
    }
    \label{fig:filtering-thresholds}
\end{figure}

\begin{CJK*}{UTF8}{gbsn}
    \begin{table*}[t]
        \centering
        \resizebox{\linewidth}{!}{
            \begin{tabular}{lp{8cm}ll}
            \toprule
            \multirow{2}{*}{\xmark} &  \textit{funny animals of the week,} & $\rightarrow$ & \textit{Animaux drôles, Animaux drôles,} \\
            & \textit{funny animal photo, cute animal pictures} & & \textit{Animaux drôles, Animaux drôles} (FRA) \\
            \multirow{2}{*}{\xmark} & \textit{damask seamless floral pattern, ornament} & $\rightarrow$ & \textit{Mifano ya Mifano ya Mifano ya Mifano ya} \\
            &  &  & \textit{Mifano ya Mifano ya Mifano ya Mifano ya Mifano ya Mifano} (SWA) \\
            \xmark  & \textit{plaid, over garment , outfit idea cute fall outfit idea} & $\rightarrow$ & \textit{方格, over garment, cute fall} (CMN) \\
            \bottomrule
            \end{tabular}
        }
    \caption{Examples of translations that are filtered out by the proposed procedure.}
    \label{t:filtered_examples}
    \end{table*}
\end{CJK*}

A commonly used dataset for multimodal pretraining is Conceptual Captions \citep{sharma-etal-2018-conceptual}, gathered from alt-text on the Internet and postprocessed to remove proper names.
We translate 2.77M English sentences from the Conceptual Captions training split into the twenty target languages in IGLUE.
Once again, we use the M2M-100-large model \citep{fan2021jmlr}, with 1.2B parameters.

We notice that the quality of the translations varies across languages, presumably due to the amount of data used to train M2M-100.
Moreover, captions in this dataset often consist of sentence fragments, which may be harder to translate well.

In order to prevent bad data from corrupting the model, we apply a filtering step to the translated data.
The two most frequent types of errors are single words being repeated multiple times and
English words being copied into the translation.
We discard sentences that exhibit these characteristics based on the following two ``badness'' scores:
\begin{itemize}
  \item \textit{Complement of the token-to-type ratio.}
  The token-to-type ratio (TTR) measures the fraction of unique tokens in a given text.
  We use its complement ($1 - \text{TTR}$),
  such that a large score (close to one) indicates repetition.
  
  \item \textit{BLEU score between the source sentence and its translation.}
  We measure the similarilty between the English source and the (non-English) target by computing the BLEU score using the NLTK toolkit \citep{bird2006acl}.
  A large score (close to one) indicates that English text has been copied into the translation.
\end{itemize}

We estimate thresholds for the two scores by manually inspecting a subset of 2{,}000 sentences from each of the twenty target languages.
We use the same TTR threshold (0.5) for all languages (since repetition is language-independent). We observe different patterns of English copying so we set different thresholds for different language groups (Figure~\ref{fig:filtering-thresholds}):
Indo-European languages with a Latin script,
all languages with a non-Latin script, and non-Indo-European languages using a Latin script.
We will discard all sentences with scores above \textit{either} threshold from the multilingual pretraining process.
Table \ref{t:filtered_examples} shows examples of translations that are filtered out by this procedure. The first two are rejected due to repeated words, the third because English words appear in the output.

\begin{figure}
    \centering
    \includegraphics[width=0.48\textwidth]{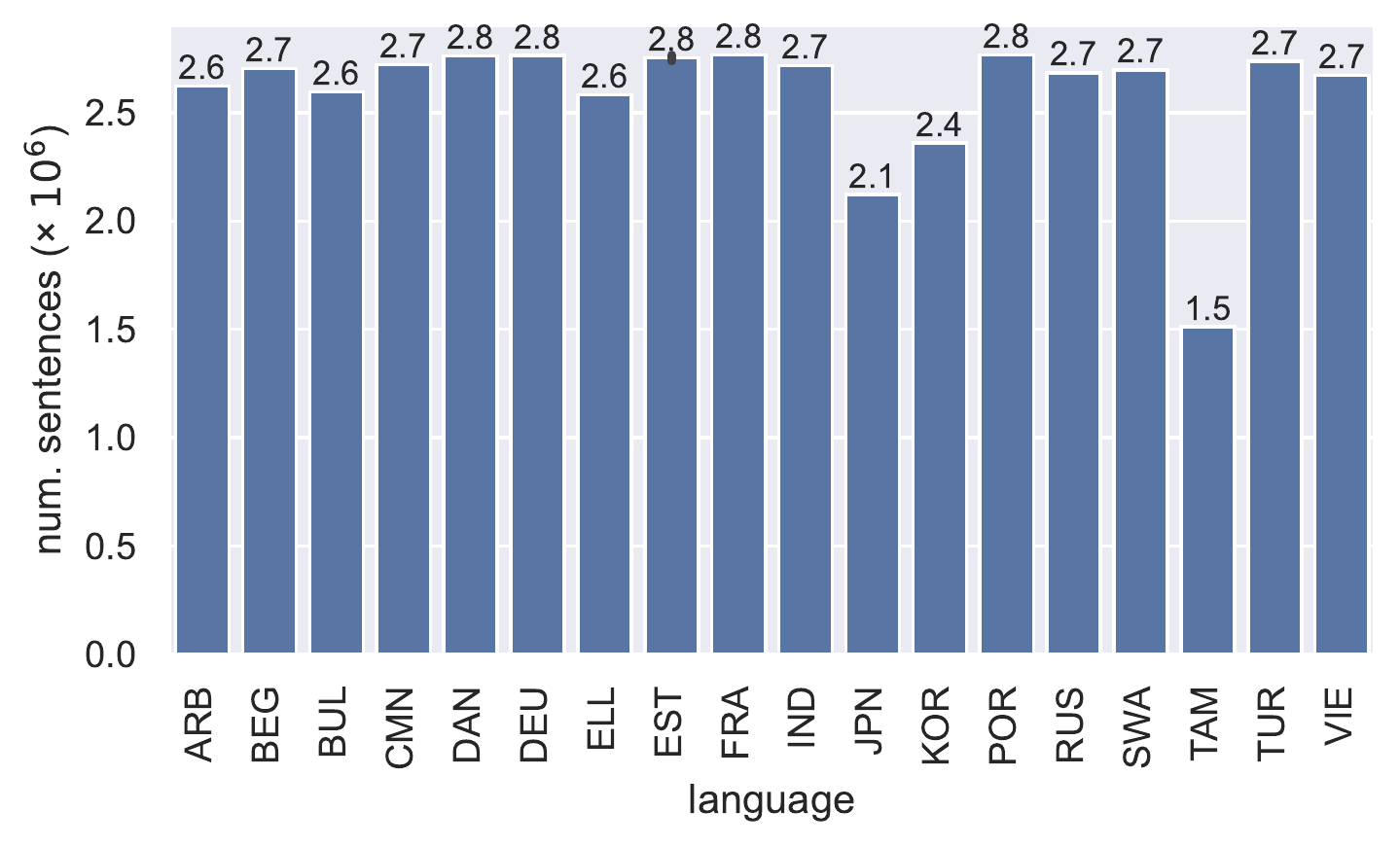}
    \caption{%
        Number of sentences (in millions) used for pretraining in each of the nineteen IGLUE languages.
        The data is obtained by translating the Conceptual Captions dataset (2.77M sentences) and filtering out the poor translations.
        The total number of sentences in the translated dataset (including English) is 52M.
    }
    \label{fig:num-sentences-after-filtering}
\end{figure}

The cumulative distribution of the scores and the corresponding thresholds are shown in Figure \ref{fig:filtering-thresholds}.
For most languages over 95\% of their translated sentences are kept,
the most notable exceptions being Tamil, Japanese and Korean, for which only 54.6\%, 76.6\%, 85.2\%, respectively, of the initial sentences are kept.
Figure \ref{fig:num-sentences-after-filtering} shows the final distribution of training data across languages.
The total number of sentences in the translated dataset for pretraining (including English) is 52M.

\section{Model}\label{sec:model}

The model we implement within our Translated Data for Multilingual Multimodal Learning framework, TD-MML,
follows the single-stream cross-modal
framework, such as UNITER \citep{chen2020eccv} and xUNITER \citep{liu-etal-2021-visually}.
It can be seen as a translate-train version of xUNITER, to which we add an additional pretraining task: visual translation language modelling~\citep{Zhou_2021_CVPR}.

The TD-MML architecture consists of a 
series of Transformer blocks, which first concatenate the visual and language 
embeddings as the input, and then passes these into a multi-layer Transformer to encode the 
contextualized representations across image-text modalities and languages.

The input sequences are image-text pairs
($\textbf{V}, \textbf{X}$), where
$\textbf{V}$ are the visual features and
$\textbf{X}$ are the embedding sequence of the corresponding caption.
The image features $\textbf{v}$ in $\textbf{V} = \{\textbf{v}_1, \textbf{v}_2, \dots, \textbf{v}_N \}$ correspond to $N=36$ object proposals extracted with Faster R-CNN \citep{NIPS2015_5638}. 
We extend the English pretraining text to also include the machine translated captions in $m$ languages:
$\left\{\textbf{x}^{l_1}, \textbf{x}^{l_2}, \dots, \textbf{x}^{l_{m}}\right\}$,
The captions are processed by the same SentencePiece tokeniser regardless of their corresponding language. 

\subsection{Pretraining Tasks}
\label{subsec:pretraining-tasks}

To learn the contextualised representation across modalities and
languages, we pretrain our model with three types of pretraining tasks 
introduced below. 
The goal of these pretraining 
tasks is to learn both the cross-modal alignment between each language and the visual modality, as well as alignments between the different languages.

\paragraph{Masked language and region modelling.}
In the V\&L pretraining literature, Masked Language Modelling (MLM) and 
Masked Region Modelling (MRM) are two mainstream pretraining tasks, which 
have been demonstrated to be effective in UNITER. 
Given the visual features 
$\textbf{V}$ and the corresponding text $\textbf{x}^l$ in 
language $l$, MLM randomly masks tokens in the text with 15\% probability and predicts
the identity of the masked token using remaining textual and visual features. 
Analogously to MLM, MRM samples and masks image regions with 15\%
probability and replaces the region input with zeros. 
The MRM task is to classify the top-1 object label of the masked 
visual feature region.
Please refer to \citet{chen2020eccv} 
for more details.

\paragraph{Image--Text Matching (ITM).}
This task attempts to determine whether an image--text pair is matched or not.
As such, this enables the model to learn the alignment of the language and vision modalities.
The matching score $s_{\theta}(\textbf{x}^l,\textbf{v})$ is computed based on the special  \texttt{[CLS]} token which is passed through a fully-connected layer with sigmoid activation function.
To train the ITM objective, we sample a positive or negative caption with equal probability for a given input image from the dataset $D$;
the selected caption is sampled uniformly from one the twenty languages in the pretraining dataset.
The loss is defined by the following equation, where we denote whether the sampled pair is a match or not by the binary label $c$:

\begin{equation}
\begin{aligned}
\mathcal{L}_{\text{ITM}}(\theta)=& - \mathbb{E}_{(\mathbf{x}^l, \mathbf{v}) \sim D} \bigg[\textit{c} \log s_{\theta}(\mathbf{x}^l, \mathbf{v}) \\
+ & (1-\textit{c})\log\left(1-s_{\theta}(\mathbf{x}^l, \mathbf{v})\right)\bigg]  \label{eq:obj_itm}
\end{aligned}
\end{equation}

\paragraph{Visual Translation Language Modeling.}
VTLM is a training objective adopted from UC$^2$ \citep{Zhou_2021_CVPR}
that combines both cross-language and cross-modal alignment learning.
It takes a triple of an image $\textbf{v}$, an English caption $\textbf{x}^{\text{ENG}}$, and a corresponding caption $\textbf{x}^l$ in a different language~$l$.
The task is to predict the masked caption tokens, using the multilingual textual input as well as the visual input.
During pretraining, we  use the same masking strategy as in MLM to randomly mask 15\% of tokens from English caption and 15\% of tokens from the other language caption.%
\footnote{This is different from \citet{Zhou_2021_CVPR}, who attempt to match the tokens across languages for co-masking. \citet{caglayan-etal-2021-cross} used the same setup as ours, where randomly mask instead of co-masking across languages.}
The loss function is:
\begin{equation}
\begin{aligned}
\mathcal{L}_{\text{VTLM}}(\theta) &= - \mathbb{E}_{(\mathbf{x}^{{\text{ENG}}},\mathbf{x}^l, \mathbf{v}) \sim D} \\
    &\log  P_{\theta}\left(\mathbf{x}^{\text{ENG}}_{a},   \mathbf{x}^{\textit{l}}_b \, | \, \mathbf{x}^{\text{ENG}}_{\setminus a}, \mathbf{x}^{l}_{ \setminus b}, \mathbf{v}\right) 
\label{eq:obj_vtlm}
\end{aligned}
\end{equation}

\section{Experimental setup}
\label{sec:experimental-setup}

The implementation of TD-MML is built in the VOLTA framework \citep{bugliarello-etal-2021-multimodal}. %
TD-MML uses the same model configuration as the xUNITER architecture \citep{liu-etal-2021-visually},
which is initialised from the XLM-R cross-lingual language model \citep{conneau-etal-2020-unsupervised}. 

\begin{figure}[t]
    \centering
    \includegraphics[width=0.46\textwidth]{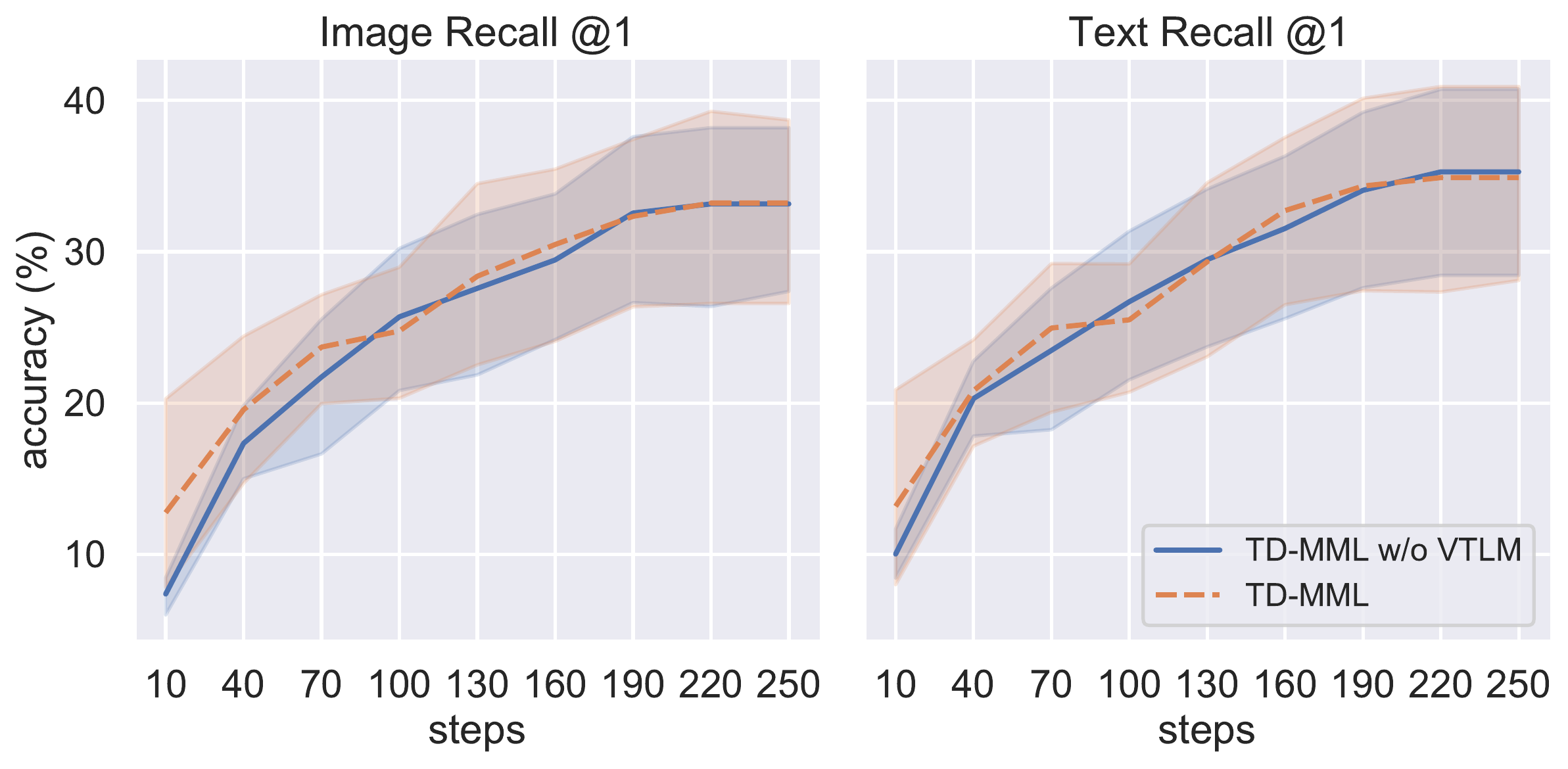}
    \caption{
    Average pretraining accuracy---image retrieval (left) and text retrieval (right)---as a function of the number of pretraining steps. Accuracy is calculated on 5K validation samples across five languages in the Conceptual Captions machine translated dataset.}
    \label{fig:evaluation}
\end{figure}

\paragraph{Pretraining.}
The dataset used for pretraining is Conceptual Captions \citep{sharma-etal-2018-conceptual}, which consists of 3.3M images with their English alt-text descriptions, of which we only have access to 2.7M instances due to linkrot.
The images are represented using ResNet-101 features extracted from $N=36$ object proposals from the Faster R-CNN~\citep{NIPS2015_5638} object detector trained on the Visual Genome dataset~\citep{8578734}.

As described in Section \ref{subsec:translation}, we translate the captions in the 19 non-English IGLUE languages with the 1.2B-parameter M2M-100-large model \citep{fan2021jmlr};
the translations are then filtered to eliminate likely translation errors. During training, we iterate over the images and uniformly at random sample the corresponding caption in one of the twenty languages, i.e. a batch contains samples from multiple languages.
We reuse the training hyperparameters of \citet{bugliarello-etal-2021-multimodal}. %
Specifically, xUNITER is trained over 2.77M image--English caption pairs, while TD-MML is pretrained on 52M image--multilingual caption pairs on 3$\times$24GB TITAN RTX for 250,000 steps, which takes 5 days.

Figure~\ref{fig:evaluation} shows the Recall@1 curves for image retrieval and text retrieval on 5K validation samples from the Conceptual Captions dataset as a function of the number of pretraining steps. The samples are chosen from languages that represent the original MaRVL languages, i.e. ENG, IND, JPN, SWA and CMN, where the non-English data comes from the machine translation process. Model performance continues to increase in both metrics until 220{,}000 updates. Therefore, we use this checkpoint to fine-tune on the downstream tasks.

\begin{table*}[th!]
	\centering
    \begin{tabular}{lccccccc}
		\toprule
		\multirow{2}{*}{\textbf{Model}} & \textbf{NLI} & \textbf{QA} & \textbf{Reasoning} & \multicolumn{4}{c}{\textbf{Retrieval}} \\
		\cmidrule(lr){2-2}\cmidrule(lr){3-3}\cmidrule(lr){4-4}
		\cmidrule(lr){5-8}
		& XVNLI & xGQA & MaRVL & \multicolumn{2}{c}{xFlickr\&CO} & \multicolumn{2}{c}{WIT} \\
		& & & & IR& TR& IR & TR \\
		\midrule
		mUNITER & 53.69 & 9.97 & 53.72 & 8.06 & 8.86 & 9.16 & 10.48\\
		xUNITER &  58.48 & 21.72 & 54.59 & 14.04 & 13.51 & 8.72 & 9.81\\
		UC$^2$ & 62.05 & 29.35 & 57.28 & 20.31 & 17.89 & 7.83 & 9.09\\
		M$^3$P & 58.25 & 28.17 & 56.00 & 12.91 & 11.90 & 8.12 & 9.98\\
		\midrule
		TD-MML & 64.84 & \textbf{35.95} & \textbf{59.67} & \textbf{21.30} & \textbf{26.35} & \textbf{9.76} & 10.35\\
  		- w/o VTLM & \textbf{66.28} & 33.01 & 58.14 & 20.90 & 24.61 & 9.14 & \textbf{10.61}\\
		\bottomrule
	\end{tabular}
	\caption{
    Average zero-shot performance on non-English languages of multimodal models for the V\&L evaluation tasks in IGLUE. Best results are marked in bold. 
The performance measure is accuracy for all the tasks except the cross-modal retrival tasks, which use Recall@1.
    }\label{Average-zero-shot}
\end{table*}

\begin{table*}[t]
	\centering
    \begin{tabular}{llc|cccccccc}
		\toprule
		\textbf{Type} & \textbf{Method} & \textbf{ENG} & \textbf{IND} & \textbf{SWA} & \textbf{TAM} & \textbf{TUR} & \textbf{CMN}  & \textbf{avg}\\
		\midrule
		\multicolumn{9}{l}{\textit{Fine-tune with English-only data (zero-shot)}} \\
		\midrule
		\multirow{2}{*}{---} & 
		xUNITER & \textbf{71.55} & 55.14 & 55.51 & 53.06 & 56.19 & 53.06 & 54.59 \\
		& TD-MML & 69.00 & 59.04 & 61.01 & 56.44 & 61.95 & 59.88 & 59.67 \\
		\midrule
	     \multicolumn{9}{l}{\textit{Fine-tune with machine translated data}} \\
		\midrule
        \multirow{2}{*}{\textit{Full}} &
		xUNITER & 67.92 & 59.57 & 61.37 & 60.39 & 64.32 & 59.39 & 61.01 \\
	     
		& TD-MML & 67.52 & 59.40 & \textbf{62.18} & 60.55 & \textbf{66.27} & 59.59 & \textbf{61.60} \\
		\midrule
        \multirow{2}{*}{\textit{Filtered}} &
		xUNITER & 67.52 & \textbf{60.82} & 61.55 & 60.63 & 63.48 & 59.88 & 61.27 \\
		
		& TD-MML & 67.09 & 57.62 & 61.91 & \textbf{61.35} & 64.58 & \textbf{60.28} & 61.15 \\
		\bottomrule
	\end{tabular}%
	\caption{
    MaRVL accuracy results for zero-shot cross-lingual evaluation, i.e. English-only NLVR2 fine-tuning, and multilingual fine-tuning using machine translated NLVR2 data (either with the full or filtered translated data). All of the models are fine-tuned for a similar number of updates.
    The average results exclude ENG accuracy.}
    \label{zero-shot-MaRVL-3-epochs}
\end{table*}

\paragraph{Fine-tuning on downstream tasks.}

We employ a translated data procedure at fine-tuning as well, using the same data filtering steps. Similar to the initial experiments in Section \ref{sec:xuniter-finetune}, we match the number of parameter updates between the experiments with English-only data and the ones with translated multilingual data. We use the same setup as in IGLUE when fine-tuning on downstream tasks.

\section{Results}\label{sec:results}

\invisible{ %
\subsection{Machine translated Fine-tune}

In order to control the quality of our machine translated multilingual training datsets, we further filtering pool translation sentences using the same filtering operation for the 
\textit{full translation}, named by \textit{filtered 
translation}. The TD-MML with English-only fine-tune 
and TD-MML machine translated fine-tune are corresponding
denoted \texttt{TD-MML} $\rightarrow$ \texttt{EN} and \texttt{TD-MML} 
$\rightarrow$ \texttt{TT}.

The accuracy results on the MaRVL task (Table 
\ref{zero-shot-MaRVL}) between \texttt{TD-MML} $\rightarrow$ 
\texttt{EN} and \texttt{TD-MML} 
$\rightarrow$ \texttt{TT} reveals that the machine translated 
fine-tune shrink the gap of new language. Besides, a 
few poorly translated sentences are removed by the filtering 
operation. The differences between filtered translations and 
full translations indicate that machine translated training and
filtering strategy lead to slightly improve performance. 
Controlling language quality is also benefit to cross-lingual 
fine-tune.

} %

\subsection{TD Pretraining and English Fine-tuning}

Here, we evaluate the zero-shot language understanding abilities of the TD-MML model that has been pretrained with multiple languages, but fine-tuned on English task-specific data only (e.g. NLVR2 for MaRVL, GQA for xGQA, etc.).
The averaged zero-shot results across languages are shown in Table 
\ref{Average-zero-shot}.
The full zero-shot per-language 
results on each task are detailed in Appendix~\ref{sec:appendix-per_lang}. 

We see a substantial improvement for TD-MML across all tasks compared to xUNITER and the state-of-the-art UC$^2$.
The improvement between our TD-MML and the best baseline models for each task reaches
4.23 points  for XVNLI,
6.6 points for xGQA,
2.39 points for MaRVL,
0.99 (IR) and 8.46 points (TR) for xFlickr\&CO, and
0.6 (IR) and 0.13 points (TR) for WIT.
The results from the other tasks show the clear benefit of pretraining on multilingual multimodal data on a diverse array of multimodal tasks across many languages.

\paragraph{Gains from VTLM.} 
Table~\ref{Average-zero-shot} also shows the effectiveness of pretraining with the additional visual translation language modeling (VTLM) objective: %
adding VTLM boosts the performance for five out of the seven tasks.
Improving cross-lingual alignment during pretraining thus manifests itself in better multi-lingual understanding ability.

\invisible{ %
\paragraph{Multilingual Models.}
From experimental results Tables in Appendix 
\ref{sec:appendix-per_lang} (\ref{zero-shot-xvnli}-\ref{zero-shot-wit}) on each task, we observe 
variance among multilingual multimodal models.
UC$^2$ is remarkably better than other 
baselines across the IGLUE benchmark (except for WIT). 
Compared to the other baseline models, the breakthrough of UC$2$ is the first attempted to pretrain multilingual 
multimodal with the translated Conceptual Caption on five languages. For the zero-shot 
performance on xFlickr\&CO (Table \ref{zero-shot-xFlickr&CO} in Appendix 
\ref{zero-shot-xGQA}), UC$^2$ achieves a large improvement on German, Janpanese, and 
Mandarin languages that have been trained on UC$^2$ if we compare the difference between
UC$^2$ and other baselines. That verified the effectiveness of machine translation captions
for multilingual transfer in V\&L pretraining model. 
} %

\subsection{MT Fine-tuning TD-MML}

\begin{table*}[t]
	\centering
    \begin{tabular}{llc|cccccccccc}
		\toprule
		\textbf{Type} & \textbf{Method} & \textbf{ENG} & \textbf{BEN} & \textbf{DEU} & \textbf{IND} & \textbf{KOR} & \textbf{POR}  & \textbf{RUS} & \textbf{CMN} & \textbf{avg}\\
		\midrule
		\multicolumn{9}{l}{\textit{Fine-tune with English-only data (zero-shot)}} \\
		\midrule
		\multirow{2}{*}{---} & 
		xUNITER & \textbf{54.83} & 10.80 & 34.83 & 33.73 & 12.12 & 22.13 & 18.84 & 19.55 & 21.70 \\
		& TD-MML & 53.60 & 23.62 & 42.29 & 38.23 & 32.44 & 41.36 & 38.08 & 35.61 & 35.95 \\
		\midrule
	     \multicolumn{9}{l}{\textit{Fine-tune with machine translated data}} \\
		\midrule
        \multirow{2}{*}{\textit{Full}} &
		xUNITER & 48.08 & 41.76 & \textbf{46.47} & \textbf{45.68} & \textbf{44.76} & \textbf{46.77} & \textbf{46.19} & \textbf{45.66} & \textbf{45.33} \\
		 
		& TD-MML & 47.38 & 42.27 & 46.31 & 44.24 & 44.53 & 46.24 & 45.78 & 44.68 & 44.86 \\
		\midrule
        \multirow{2}{*}{\textit{Filtered}} &
		xUNITER & 48.40 & 42.10 & 46.13 & 45.30 & 44.70 & 46.33 & 45.95 & 45.50 & 45.14\\
		 
		& TD-MML & 48.04 & \textbf{42.86} & 46.45 & 44.78 & 44.60 & 46.67 & 46.02 & 45.07 & 45.21\\
		\bottomrule
	\end{tabular}%
	\caption{%
        xGQA accuracy results for
        English-only fine-tuning (zero-shot evaluation) and 
        multilingual fine-tuning using machine translated GQA data (either with the full or filtered translated data). All of the models are fine-tuned for a similar number of updates.
        The average results exclude ENG accuracy.
    } %
    
    \label{zero-shot-xGQA-1-epoch}
\end{table*}

\begin{figure*}[t]
    \centering
    \includegraphics[width=\textwidth]{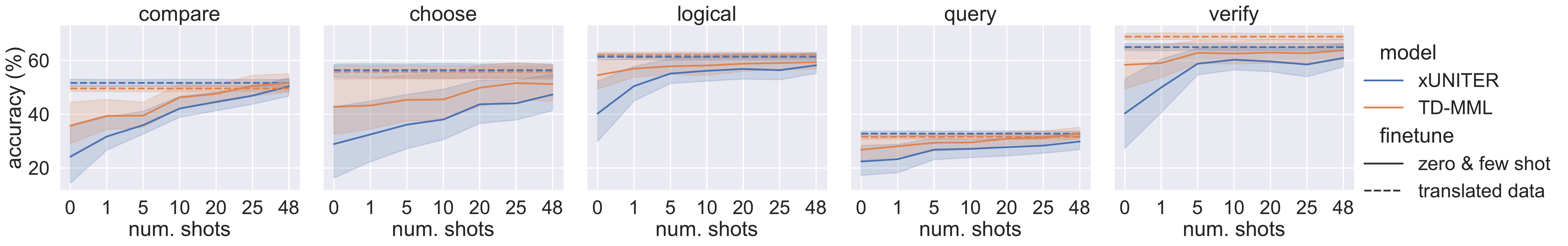}
    \caption{xGQA average accuracy (with confidence intervals) across the languages on the five question types. xUNITER and TD-MML are evaluated in the zero-shot, few-shot and machine translation fine-tuning settings.
    The error bars represent 95\% confidence intervals and
    are obtained by bootstraping (1000 repeats).
    \label{fig:xgqa-few-shot}
    }
\end{figure*}

We now ask whether combining the machine translated pretraining strategy of TD-MML with additional machine translated fine-tuning can provide further gains, compared to both English-only fine-tuning (zero-shot) and the MT-fine-tuning strategy applied to xUNITER in Section~\ref{sec:xuniter-finetune}.
The results for MaRVL and xGQA are shown in Table \ref{zero-shot-MaRVL-3-epochs} and Table \ref{zero-shot-xGQA-1-epoch} respectively.
Perhaps surprisingly, the performance for xUNITER and TD-MML are very similar after multilingual MT fine-tuning, with TD-MML slightly outperforming xUNITER on the MaRVL dataset
and with xUNITER generally yielding better performance on xGQA.
That is, in this setting, multilingual multimodal pretraining in TD-MML does not convey any added benefit. A potential explanation is that fine-tuning on enough multilingual data (as it is the case of machine translated data) compensates for the lack of multilingual multimodal pre-training. The performance of TD-MML on the English splits after English-only fine-tuning supports this hypothesis.

The results in Tables \ref{zero-shot-MaRVL-3-epochs} and \ref{zero-shot-xGQA-1-epoch} further indicate that the filtering strategies offer mixed results when applied to fine-tuning.
We believe that this outcome happens because the corresponding datasets, GQA for xGQA and NLVR2 for MaRVL, are typically much cleaner datasets than the Conceptual Captions dataset, resulting in less filtering and, consequently, less representative results.

\subsection{Few-Shot vs Machine Translated Data}

\begin{figure*}
    \centering
    \footnotesize
    \newcommand\mylabel[1]{\textsf{\scriptsize \color{gray}#1}}
    \setlength{\tabcolsep}{4pt}
    \begin{tabular}{ccc|cc|cc|cc}
               & \multicolumn{2}{c}{\mylabel{Q1}}
               & \multicolumn{2}{c}{\mylabel{Q2}}
               & \multicolumn{2}{c}{\mylabel{Q3}}
               & \multicolumn{2}{c}{\mylabel{Q4}}
               \\
               & \multicolumn{2}{c}{\includegraphics[height=2.2cm]{imgs/xgqa-images/n393305.jpg}}
               & \multicolumn{2}{c}{\includegraphics[height=2.2cm]{imgs/xgqa-images/n83784.jpg}} 
               & \multicolumn{2}{c}{\includegraphics[height=2.2cm]{imgs/xgqa-images/n470920-dpi72.jpg}}
               & \multicolumn{2}{c}{\includegraphics[height=2.2cm]{imgs/xgqa-images/n379991.jpg}}
               \\
               \mylabel{Q}
               & \multicolumn{2}{c|}{What is the sign behind} %
               & \multicolumn{2}{c|}{Is the black and white}                 
               & \multicolumn{2}{c|}{What is covering the man}  %
               & \multicolumn{2}{c}{How is this cooking}
               \\
               & \multicolumn{2}{c|}{the young person?}
               & \multicolumn{2}{c|}{cat unhappy or happy?}                 
               & \multicolumn{2}{c|}{that is wearing jeans?}
               & \multicolumn{2}{c}{utensil called?}
               \\
               \mylabel{A}
               & \multicolumn{2}{c|}{traffic sign}
               & \multicolumn{2}{c|}{unhappy}
               & \multicolumn{2}{c|}{jacket}
               & \multicolumn{2}{c}{baking pan}
               \\
            \hline
               & \mylabel{fine-tune: ENG} & \mylabel{fine-tune: MT}
               & \mylabel{fine-tune: ENG} & \mylabel{fine-tune: MT}
               & \mylabel{fine-tune: ENG} & \mylabel{fine-tune: MT}
               & \mylabel{fine-tune: ENG} & \mylabel{fine-tune: MT}
               \\
            \mylabel{BEN}  &          car &       pole &       no &   lush &      bag &  umbrella &  paper &     pretzel \\
            \mylabel{CMN}  &         pole &  stop sign &    white &  happy &     coat &  umbrella &   book &       stove \\
            \mylabel{DEU}  & fire hydrant &  stop sign &  unhappy &  happy &   jacket &  umbrella & mirror &  tea kettle \\
            \mylabel{ENG}  &    stop sign &  stop sign &  unhappy &  happy &    towel &  umbrella &    pan &  tea kettle \\
            \mylabel{IND}  &  street sign &  stop sign &  unhappy &  happy &  blanket &  umbrella &    yes &         yes \\
            \mylabel{KOR}  &         pole &  stop sign &     gray &  happy & backpack &  umbrella &   drum &        drum \\
            \mylabel{POR}  &          car &        car &    happy &  happy & suitcase &  umbrella &  table &  tea kettle \\
            \mylabel{RUS}  &    stop sign &  stop sign &    happy &  happy & umbrella &  umbrella &  shelf &         pan \\
        \hline
    \end{tabular}
    \caption{%
        Qualitative results on the xGQA dataset.
        Given an image and a question (denoted by Q), we show the corresponding groundtruth answer (denoted by A), together with the predictions in each of the eight languages for the model finetuned
        on either English-only sentences (left column of answers) or
        on machine translated sentences (right column of answers).
    }
    \label{fig:qualitative-xgqa}
\end{figure*}

We also ask whether it is better for downstream performance
to train on a limited number of clean language-specific and task-specific samples (few-shot learning) 
or to simply machine translate the task-specific English data, which is likely to result in noisier data.
So, in addition to the previously introduced fine-tuning setups---%
fine-tuning on task-specific English data for a zero-shot evaluation setting, and
fine-tune on machine translated task-specific English data---%
we also consider the few-shot learning setup,
in which we continue fine-tuning of the zero-shot (English fine-tuned) model, using a few human-authored language-specific and downstream-specific samples  from the IGLUE benchmark (i.e., 1, 5, 10, 20, 25, 48 shots).

Few shot results for xGQA are presented in Figure \ref{fig:xgqa-few-shot}, broken down by question type.
We observe that, as expected, the performance generally improves with the quantity of training data (number of shots).
More interestingly, %
the machine translated fine-tuning upper bounds the performance of the few-shot approach for both of our multimodal multilingual models.
Across the five question categories, the variation in performance can be largely explained by the cardinality of the set of plausible answers: it is easier to answer correctly a verification or logical question, whose answer is usually either "yes" or "no", than a querying question, whose answer usually involves a broader set of words.

\subsection{Cross-Language Correlation Analysis}

\begin{figure}[t]
    \centering
    \includegraphics[width=0.23\textwidth]{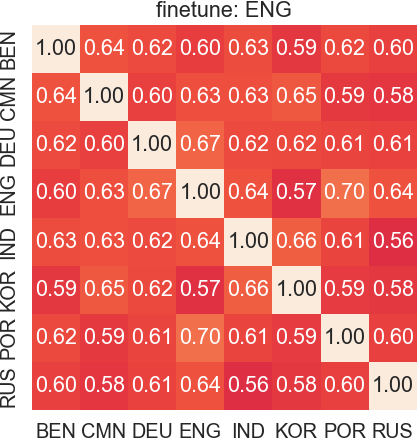}%
    \hspace{0.1em}
    \includegraphics[width=0.23\textwidth]{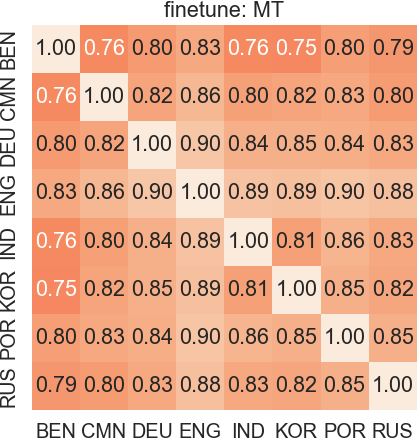}
    \caption{%
        Cross-language prediction correlations on the test split of xGQA for two of the proposed models:
        TD-MML fine-tuned on English-only data (left) and
        TD-MML fine-tuned on translated data (right).
    }
    \label{fig:cross-language-correlation}
\end{figure}

Are the same questions easy (answered correctly) or difficult (answered incorrectly) across languages?
We use Cohen's kappa coefficient $\kappa$ to measure agreement between languages on the xGQA dataset (see Appendix \ref{kappa_setting} for details). The results are presented in Figure \ref{fig:cross-language-correlation}.
We show results for two variants of our pretrained TD-MML model:
either fine-tuned on English-only data (ENG) or on machine translated data (MT).

We see that the MT fine-tuned results show much higher agreement across languages, compared to English-only fine-tuning.
On the one hand, this could be considered counter intuitive: in the MT fine-tuned setting, there is more language-specific data.
However, the MT fine-tuned results have higher accuracy overall, suggesting that high agreement across languages is due to the model confronting inherent item difficulty (as judged by all languages), rather than language-specific issues.

Examples of increased cross-language agreement in the MT fine-tuned model (MT) are presented in Figure \ref{fig:qualitative-xgqa}.
Across the eight languages, we find the predictions of the MT fine-tuned model are more consistent than those of the ENG-finetuned model (Q1, Q2, Q3).
However, for more ambiguous and difficult samples, the model fine-tuned with translated data still gives varied, but arguably more plausible, predictions across languages (Q4).

\invisible{ %
\begin{figure*}
    \centering
    \includegraphics[width=0.32\textwidth]{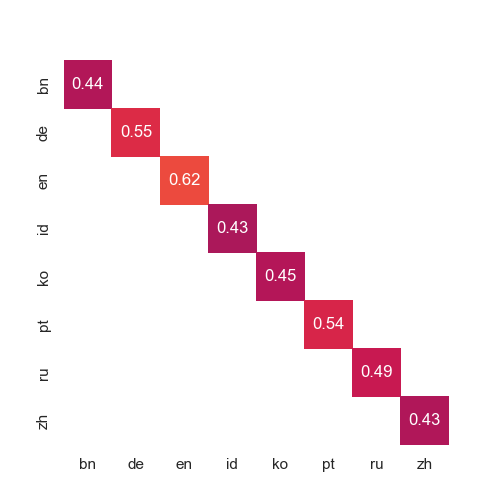}
    \includegraphics[width=0.47\textwidth]{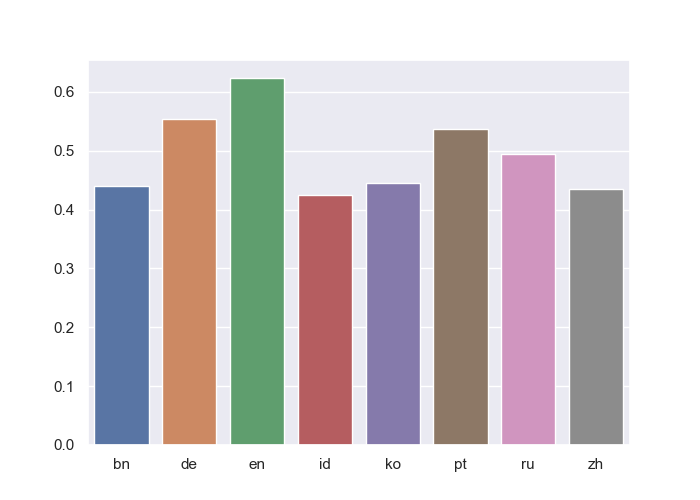}
    \caption{%
        Cross-model prediction correlations on the test split of the xGQA dataset between two of the proposed models:
        TD-MML finetuned on English-only data  TD-MML finetuned on translated data.
        We measure the correlation as Cohen's Kappa ratio ($\kappa$ normalized by the maximum achievable value).
    }
    \label{fig:cross-model-correlation}
\end{figure*}

}

\section{Conclusion}
\label{sec:conclusions}

In this paper, we investigate the role of machine translated (MT) data in multilingual multimodal learning in a controlled setup.
We consider two applications of MT data, namely for augmenting the pretraining and the fine-tuning data.
We find that both convey a clear immediate benefit on downstream performance for nearly all tasks in the IGLUE benchmark; however, we do not find an additive benefit of combining it for both pretraining and fine-tuning together.
When using machine translated text, filtering out bad translations in a quick and reliable way is crucial, and we develop a simple and effective strategy for doing this.
Our results shed light in the importance of explicitly grounding multilingual text in the visual modality in both pretraining and fine-tuning stages.

\section*{Limitations}

Our paper investigates the benefits and limitations of machine translated data towards multilingual multimodal learning.
In doing so, we solely rely on the M2M-100 model~\citep{fan2021jmlr}. This is a large, multi-to-multi translation system, which proved to be easy to use.
Our analyses and results are based on the performance of this model.
It would be instructive to investigate how the expected performance of translation systems\footnote{Despite our best efforts, we were unable to find the per-language translation performance for the M2M-100 model.} affects (i) the proportion of sentences with high `badness' scores, and (ii) the resulting performance of the multilingual multimodal systems. %
Moreover, while machine translating a large corpus is a cheaper effort than manually translating the data or scraping it from the web, there is still a one-time effort required to translate the data before using it for training new models. Therefore, we release our multilingual pretraining and fine-tuning datasets.

From an experimental angle, although the proposed framework can be applied to any existing architecture, we only evaluate a single model due to computational constraints.

We would also like to stress the importance of using target-language originating evaluation data in multimodal setups, rather than translated data.
Fitting to translationese is a risk when using translation data at training time, and can only be identified if the evaluation data does not also contain translations, especially automatically generated ones.

Finally, a core limitation of the overall translate data framework is that it centers English as the source language.
For example, this means only concepts mentioned in English captions can be grounded across languages~\citep{liu-etal-2021-visually}, and hence some non-English concepts might never be modelled.
However, we show that machine translating data provides a strong starting point that can effortlessly be integrated in a pipeline, upon which language-specific annotations can be added.

\section*{Acknowledgements}
We thank the anonymous reviewers for their constructive feedback.
This work was supported in part by a grant of the Romanian Ministry of Education and Research No PN-III-P1-1.1-PD-2019-0918 (Dan Oneață).
{\scriptsize\euflag} Emanuele Bugliarello was supported by the funding from the European Union's Horizon 2020 research and innovation programme under the Marie Sk\l{}odowska-Curie grant agreement No 801199.

\bibliography{anthology,custom,iglue}
\bibliographystyle{acl_natbib}

\clearpage
\appendix

\section{Language information}
\label{sec:appendix-language}

\begin{table*}
 	\centering
 	\small
     \begin{tabular}{lcccccccc}
 		\toprule
 		\multicolumn{4}{c}{\textbf{Language}} & \textbf{NLI} & \textbf{QA} & \textbf{Reasoning} & \multicolumn{2}{c}{\textbf{Retrieval}} \\
 		\cmidrule(lr){1-4}\cmidrule(lr){5-5}\cmidrule(lr){6-6}\cmidrule(lr){7-7}\cmidrule(lr){8-9}
 		Name & Code & Family & Script & XVNLI & xGQA & MaRVL & xFlickr\&CO & WIT \\
 		\midrule
         English & ENG & Indo-E & Latin  & \doublecheck & \doublecheck & \doublecheck & \doublecheck & \doublecheck   \\
         \midrule
         Arabic & ARB & Afro-A & Arabic  & \doublecheck & & & & \checkmark \\
         Bengali & BEN & Indo-E & Bengali  & & \doublecheck & & &  \\
         Bulgarian & BUL & Indo-E & Cyrillic  & & & & & \checkmark  \\
         Danish & DAN & Indo-E & Latin  & & & & & \checkmark  \\
         Estonian & EST & Uralic & Latin &  & & & & \checkmark  \\
         German & DEU & Indo-E & Latin &  & \doublecheck& & \doublecheck & \\
         Greek & ELL & Indo-E & Greek  &  & & & & \checkmark  \\ 
         French & FRA & Indo-E & Latin &  \doublecheck & & & & \\
         Indonesian & IND & Austron & Latin  & & \doublecheck & \doublecheck  & \doublecheck &\checkmark \\
         Japanese & JPN & Japonic & Kanji &  & & &  $\circ$ \doublecheck & \checkmark \\
         Korean & KOR & Koreanic & Hangul   & & \doublecheck & &   & \checkmark \\
         Mandarin & CMN & Sino-T & Hanzi & & \doublecheck  & \doublecheck  & \doublecheck  &  \\
         Portuguese & POR & Indo-E & Latin  & & \doublecheck  &  & &     \\ 
         Russian & RUS & Indo-E & Cyrillic  & \doublecheck & \doublecheck &  & \doublecheck    &  \\ 
         Spanish & SPA & Indo-E & Latin &   \doublecheck & &  & \doublecheck &     \\ 
         Swahili & SWA & Niger-C & Latin &  & & \checkmark  &  &         \\ 
         Tamil & TAM & Dravidian & Tamil &  & & \checkmark &  &         \\ 
         Turkish & TUR & Turkic & Latin  &    &  & \doublecheck  & \doublecheck  & \checkmark  \\ 
         Vietnamese & VIE & Austro-A & Latin    & &  & &  & \checkmark  \\ 
 		\bottomrule
 	\end{tabular}
 	\caption[IGLUE details]{IGLUE details: Table replicated from Table 2 in \protect\citet{bugliarello2022}. 
 	Tasks legend: \doublecheck few-shot train and test splits; \checkmark test-only; $\circ$ Japanese captions in xFilickr\&CO are manual translations of the English ones.
     }\label{tab:IGLUE}
 \end{table*}

The machine translated fine-tuning data and pre-training data cover 20 languages,
spanning 11 language families and 9 scripts.
The scripts are Arabic, Bengali-Assamese, Chinese, Cyrillic, Greek, Huangual, Kanji, Latin, and Tamil.
Table \ref{tab:IGLUE} summarizes this information together with a listing of the language code abbreviations and details regarding the use of the languages in each of the five tasks from the IGLUE benchmark.

\section{Kappa Setting}
\label{kappa_setting}

Cohen's kappa corrects for random agreement, but has an upper bound based on the difference in marginal probabilities (here, equivalent to accuracy):
if one system assigns more correct answers than the other, then the maximum achievable $\kappa$ is less than one. Since we compare across languages with different accuracies, we normalise the usual coefficient by the maximum achievable value, resulting in $\kappa$-ratio, the proportion of agreement given system accuracies (label rates).

\section{Performance per Target Language}
\label{sec:appendix-per_lang}

\begin{table*}
	\centering
	\small
    \begin{tabular}{lccccccccc}
		\toprule
		\textbf{Method} & \textbf{ENG} & \textbf{IND} & \textbf{SWA} & \textbf{TAM} & \textbf{TUR} & \textbf{CMN} & \textbf{avg}\\
		\midrule
		mUNITER & \textbf{71.91} & 54.79 & 51.17 & 52.66 & 54.66 & 55.34 & 56.76 \\
		xUNITER & 71.55 & 55.14 & 55.51 & 53.06 & 56.19 & 53.06 &  57.42\\
		UC$^2$ &  70.56 & 56.74 & 52.62 & \textbf{60.47} & 56.70 & \textbf{59.88} &  59.50\\ 
		M$^3$P & 68.22 & 56.47 & 55.69 & 56.04 & 56.78 & 55.04 &  58.04\\
		\midrule
		TD-MML w/o VTLM & 68.45 & 58.25 & 59.30 & 56.28 & 61.02 & 55.83 & 59.85 \\   
		TD-MML & 69.00 & \textbf{59.04} & \textbf{61.01} & 56.44 & \textbf{61.95} & \textbf{59.88} & \textbf{61.22} \\
		\bottomrule
	\end{tabular}
	\caption{
    Accuracy on the MaRVL task for zero-shot evaluation (i.e. English-only fine-tuning). 
    }
    \label{zero-shot-marvl}
\end{table*}
\begin{table*}
	\centering
	\small
    \begin{tabular}{lcccccc}
		\toprule
		\textbf{Method} & \textbf{ENG} & \textbf{ARB} & \textbf{SPA} & \textbf{FRA} & \textbf{RUS} & \textbf{avg}\\
		\midrule
		mUNITER & 76.38 & 46.73 & 56.96 & 59.36 & 51.72 & 58.23 \\
		xUNITER & 75.77 & 51.98 & 58.94 & 63.32 & 59.71 & 61.94 \\
		UC$^2$ & 76.38 & 56.19 & 57.47 & \textbf{69.67} & 64.86 &  64.91\\
		M$^3$P & \textbf{76.89} & 55.24 & 58.85 & 56.36 & 62.54 &  61.98\\
		\midrule
		TD-MML w/o VTLM & 76.12 & \textbf{62.20} & \textbf{66.75} & 68.39 & \textbf{67.78} & \textbf{68.25}\\
		TD-MML & 75.52 & 61.25 & 65.89 & 67.44 & 64.78 & 66.98 \\ 
		\bottomrule
	\end{tabular}
	\caption{
    Accuracy on the XVNLI task for zero-shot evaluation (i.e. English-only fine-tuning).
    }
    \label{zero-shot-xvnli}
\end{table*}
\begin{table*}
	\centering
	\small
    \begin{tabular}{lccccccccc}
		\toprule
		\textbf{Method} & \textbf{ENG} & \textbf{BEN} & \textbf{DEU} & \textbf{IND} & \textbf{KOR} & \textbf{POR} & \textbf{RUS} & \textbf{CMN} & \textbf{avg}\\
		\midrule
		mUNITER & 54.68 & 3.06 & 23.95 & 9.36 & 4.21 & 13.67 & 8.49 & 7.30 &  16.77 \\
		xUNITER & 54.83 & 10.80 & 34.83 & 33.73 & 12.12 & 22.13 & 18.84 & 19.55 &  26.75\\
		UC$^2$ & \textbf{55.19} & 19.99 & \textbf{42.85} & 28.67 & 21.36 & 30.42 & 31.00 & 31.16 &  32.78\\ 
		M$^3$P &53.75 & 18.64 & 33.42 & 32.48 & 25.11 & 31.40 & 27.50 & 28.65 & 31.76 \\
		\midrule
		TD-MML w/o VTLM & 54.37 & 16.20 & 39.98 & 36.28 & 29.96 & \textbf{41.79} & 37.00 & 29.88 & 35.68 \\
		TD-MML & 53.60 & \textbf{23.62} & 42.29 & \textbf{38.23} & \textbf{32.44} & 41.36 & \textbf{38.08} & \textbf{35.61} & \textbf{38.15} \\
		\bottomrule
	\end{tabular}
	\caption{
    Accuracy on the xGQA task for zero-shot evaluation (i.e. English-only fine-tuning). 
    }
    \label{zero-shot-xGQA}
\end{table*}
\begin{table*}
	\centering
	\small
    \begin{tabular}{l|lccccccccc}
		\toprule
	 	Type & \textbf{Method} & \textbf{ENG} & \textbf{DEU} & \textbf{SPA} & \textbf{IDN} & \textbf{JPN} & \textbf{RUS} & \textbf{TUR} & \textbf{CMN}  & \textbf{avg}\\
		\midrule
		\multirow{6}{*}{IR} & 
		mUNITER & \textbf{44.50} & 12.05 & 13.15 & 5.95 & 6.30 & 5.85 & 1.75 & 11.35 & 12.61\\
		& xUNITER & 38.45 & 14.55 & 16.10 & 16.50 & 10.25 & 15.90 & 9.05 & 15.95 & 17.09  \\
		& UC$^2$ & 37.40 & \textbf{28.60} & 15.95 & 14.60 & \textbf{24.25} & 20.00 & 7.15 & \textbf{31.60} & \textbf{22.44} \\
		& M$^3$P & 31.35 & 13.35 & 13.40 & 13.20 & 10.30 & 15.95 & 7.15 & 16.45 & 15.14 \\
		\cmidrule(lr){2-11}
		& TD-MML w/o VTLM & 29.45 & 19.70 & \textbf{23.20} & 21.05 & 16.65 & \textbf{24.15} & 19.65 & 21.90 & 21.97  \\
		& TD-MML &  28.15 & 19.95 & 22.20 & \textbf{22.35} & 18.35 & 23.30 & \textbf{19.75} & 23.20 & 22.16 \\
		\hline
        \hline
	   \multirow{6}{*}{TR} 
	    & mUNITER & \textbf{40.90} & 11.85 & 13.05 & 7.55 & 7.70 & 6.80 & 3.25 & 11.85 & 12.87 \\
		& xUNITER &  32.05 & 13.25 & 15.10 & 16.75 & 9.85 & 14.80 & 10.05 & 14.80 & 15.83 \\
		& UC$^2$ & 34.55 & 23.90 & 15.30 & 13.60 & 22.40 & 16.75 & 6.95 & 26.30 & 19.97 \\
		& M$^3$P & 24.60 & 11.85 & 12.15 & 12.10 & 9.65 & 14.45 & 8.35 & 14.75 & 13.49 \\
		\cmidrule(lr){2-11}
		& TD-MML w/o VTLM &  34.35 & 22.30 & 27.80 & 24.35 & 21.00 & 27.60 & 22.30 & 26.90 & 25.83 \\
		& TD-MML &  33.70 & \textbf{24.70} & \textbf{29.40} & \textbf{25.55} & \textbf{22.90} & \textbf{29.95} & \textbf{23.65} & \textbf{28.30} & \textbf{27.27}  \\
		\bottomrule
	\end{tabular}
	\caption{
    Experimental results on the xFlickr\&CO task (image retrieval, IR, top, and text retrieval, TR, bottom) for zero-shot evaluation (i.e. English-only fine-tuning). 
    }
    \label{zero-shot-xFlickr&CO}
\end{table*}

\begin{table*}
	\centering
	\footnotesize
	\setlength{\tabcolsep}{4pt}
    \begin{tabular}{l|lcccccccccccc}
		\toprule
	 	Type & \textbf{Method} & \textbf{ARB} & \textbf{BUL} & \textbf{DAN} & \textbf{ELL} & \textbf{ENG} & \textbf{EST} & \textbf{IND} & \textbf{JPN} & \textbf{KOR} & \textbf{TUR} & \textbf{VIE}   & \textbf{avg} \\
		\midrule
		\multirow{6}{*}{IR} & 
		mUNITER & 7.74 & 8.26 & 10.66 & 8.95 & \textbf{19.90} & 7.67 & 10.88 & \textbf{9.00} & 5.91 & 9.57 & \textbf{13.00} & 10.14 \\
		& xUNITER & 7.63 & 8.49 & 10.32 & 11.23 & 16.70 & 6.41 & 10.21 & 7.30 & \textbf{6.34} & 9.57 & 9.72 & 9.45 \\
		& UC$^2$ & 6.62 & 8.84 & 9.43 & 8.77 & 17.90 & 4.69 & 9.88 & 9.80 & 4.30 & 7.49 & 8.46 & 8.74 \\
		& M$^3$P & 8.87 & 8.84 & 9.43 & 9.65 & 15.50 & 5.38 & 8.66 & 7.00 & 6.12 & 6.52 & 10.78 & 8.95 \\
		\cmidrule(lr){2-14}
		& TD-MML w/o VTLM & \textbf{9.20} & \textbf{9.19} & 9.43 & 11.05 & 14.70 & 7.90 & 10.43 & 8.10 & 5.69 & 9.29 & 11.10 & 9.64 \\
		& TD-MML &  8.64 & 8.02 & \textbf{11.00} & \textbf{11.58} & 16.10 & \textbf{8.12} & \textbf{11.77} & \textbf{9.00} & 6.02 & \textbf{11.79} & 11.63 & \textbf{10.33} \\
		\hline
        \hline
	   \multirow{6}{*}{TR} 
	    & mUNITER & 9.21 & 10.17 & 12.16 & 10.54 & \textbf{22.34} & 8.33 & \textbf{12.88} & 8.79 & 6.75 & 10.87 & \textbf{15.07} & \textbf{11.56} \\
		& xUNITER &  9.08 & 10.30 & 9.34 & 12.38 & 18.54 & 7.82 & 10.66 & 10.10 & 6.97 & 9.69 & 11.74 & 10.60 \\
		& UC$^2$ & 8.32 & 7.69 & 10.44 & 11.64 & 19.71 & 6.03 & 11.47 & \textbf{10.81} & 5.74 & 8.81 & 9.90 & 10.05 \\
		& M$^3$P & 8.32 & 9.80 & 11.79 & 12.02 & 15.33 & 8.21 & 10.89 & 8.43 & \textbf{7.09} & 10.57 & 12.66 & 10.46 \\
		\cmidrule(lr){2-14}
		& TD-MML w/o VTLM &  \textbf{10.47} & \textbf{11.04} & 10.69 & \textbf{13.86} & 18.54 & \textbf{9.49} & 11.59 & 8.91 & 6.79 & 11.01 & 13.02 & 11.33 \\
		& TD-MML &  9.21 & 10.17 & \textbf{13.39} & 12.20 & 17.81 & 7.56 & 10.77 & 9.50 & 6.41 & \textbf{11.16} & 13.58 & 11.02  \\
		\bottomrule
	\end{tabular}
	\caption{\label{zero-shot-wit}
    Experimental results on the WIT tasks (image retrieval, IR, top, and text retrieval, TR, bottom) for zero-shot evaluation (i.e. English-only fine-tuning). 
    }
\end{table*}

Tables \ref{zero-shot-marvl} to \ref{zero-shot-wit} provide language-specific performance for zero-shot evaluation (i.e., English-only fine-tuning) on the five IGLUE tasks (MaRVL, XVNLI, xGQA, xFlickr\&CO, WIT)
for two variants of our TD-MML approach (with and without VTML loss) against four state-of-the-art approaches (mUNITER, xUNITER, UC$^2$, M$^3$P).

The experimental results show that our TD-MML usually achieves better performance than the competing models on the non-English languages.
The closest competitor is UC$^2$, which is also a method that is pretrained on machine translated data, but only in five languages (CMN, DEU, JPN, FRA, CZE).
This partially explains UC$^2$ strong performance in some of these instances:
for example, on FRA in XVNLI (Table \ref{zero-shot-xvnli}) or on DEU in xGQA (Table \ref{zero-shot-xGQA}).

On the WIT retrieval tasks, we notice that even if TD-MML still performs better or on par with previous approaches,
the results are poor across all languages and methods.
A possible explanation is that the distribution of images and captions on Wikipedia is distinct from other datasets. 

Among the two variants of TD-MML, we generally observe a benefit of incorporating the VTML loss,
the largest gains manifesting for BEN, CMN, KOR, RUS languages on the xGQA task. 

\section{Analysis over Question Types in xGQA}
\label{sec:appendix-type_xGQA}

\begin{figure*}
    \centering
    \begin{subfigure}[t]{0.78\textwidth}
        \includegraphics[width=\textwidth]{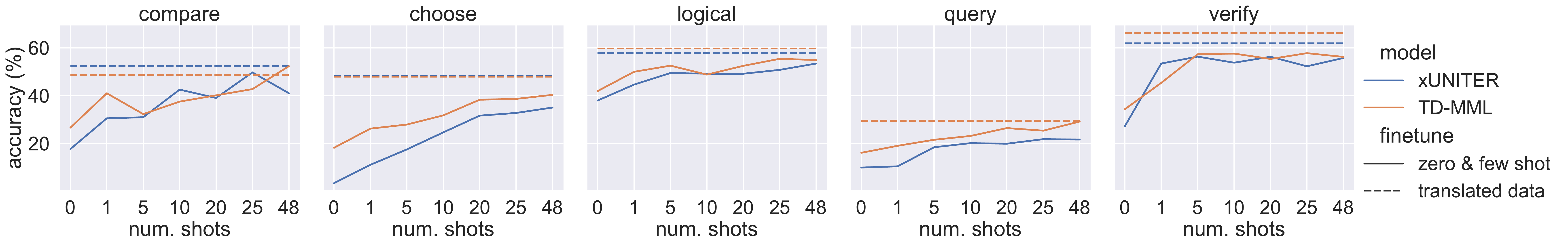}
        \subcaption{xGQA-BEN} 
    \end{subfigure}
    \begin{subfigure}[t]{0.78\textwidth}
        \includegraphics[width=\textwidth]{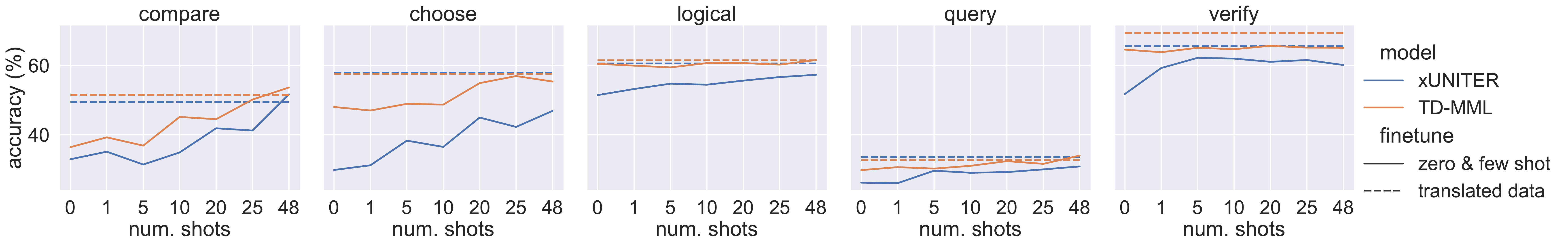}
        \subcaption{xGQA-DEU}
    \end{subfigure} 
    \begin{subfigure}[t]{0.78\textwidth}
        \includegraphics[width=\textwidth]{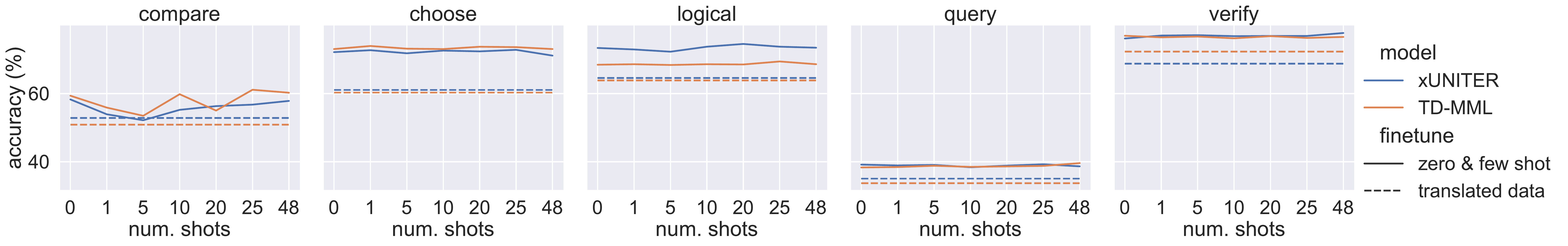}
        \subcaption{xGQA-ENG}
    \end{subfigure}
    \begin{subfigure}[t]{0.78\textwidth}
        \includegraphics[width=\textwidth]{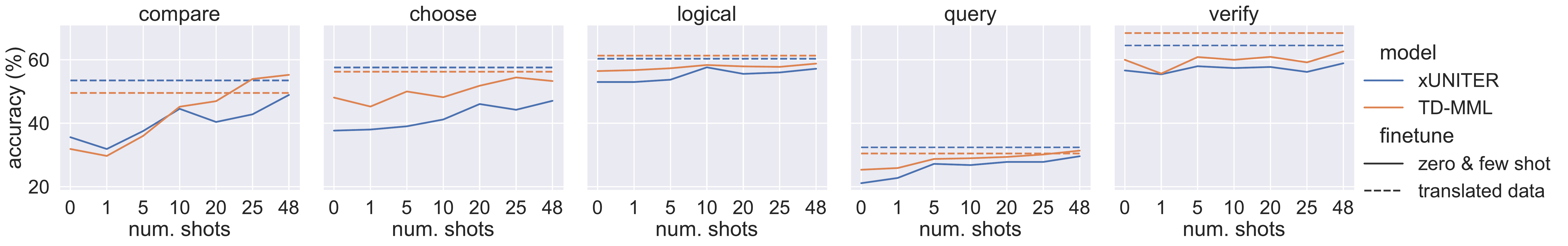}
        \subcaption{xGQA-IND}
    \end{subfigure}
    \begin{subfigure}[t]{0.78\textwidth}
        \includegraphics[width=\textwidth]{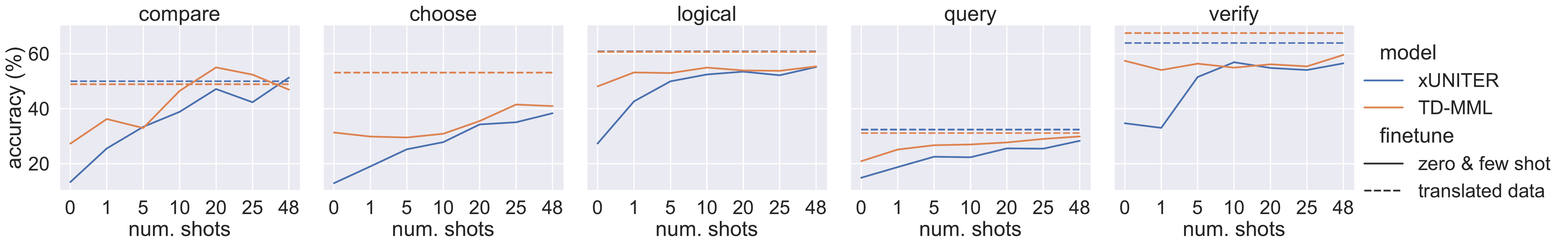}
        \subcaption{xGQA-KOR}
    \end{subfigure}
    \begin{subfigure}[t]{0.78\textwidth}
        \includegraphics[width=\textwidth]{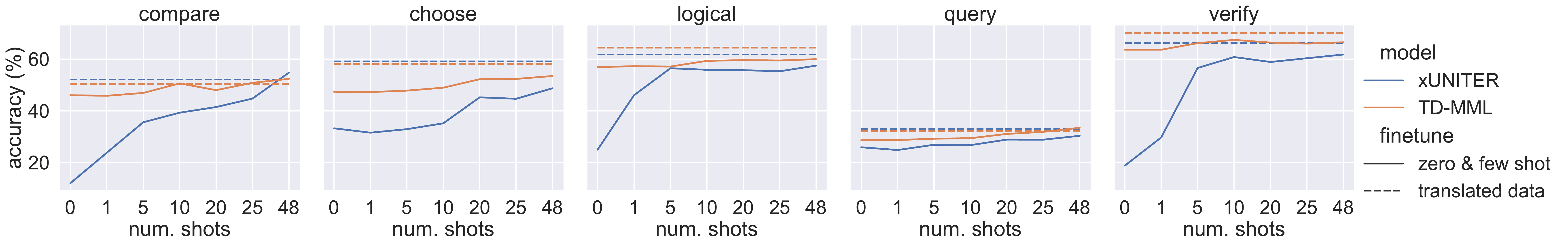}
        \subcaption{xGQA-POR}
    \end{subfigure}
    \begin{subfigure}[t]{0.78\textwidth}
        \includegraphics[width=\textwidth]{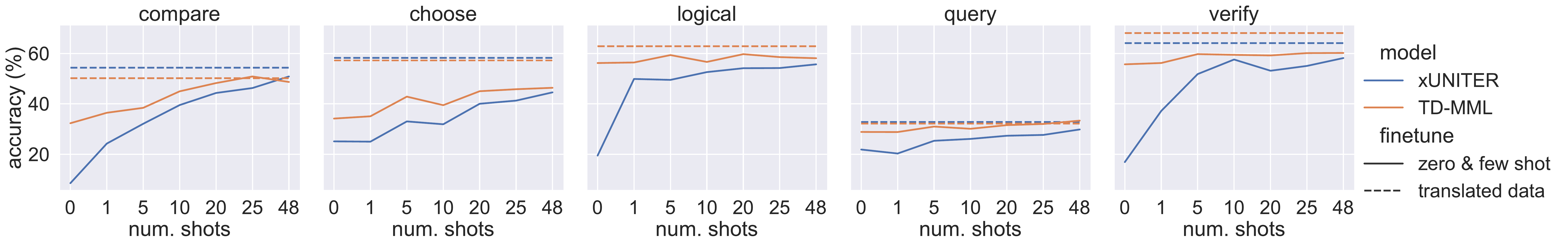}
        \subcaption{xGQA-RUS}
    \end{subfigure}
    \begin{subfigure}[t]{0.78\textwidth}
        \flushright
        \includegraphics[width=\textwidth]{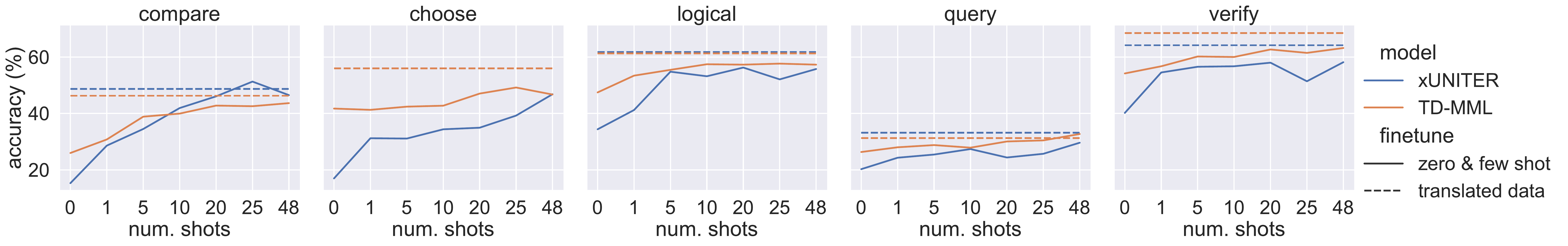}
        \subcaption{xGQA-CMN}
    \end{subfigure}
    \caption{
    xGQA accuracy on zero-shot, few shot learning and machine translation fine-tuning evaluation for each of the five questions types (compare, choose, logical, query, verify) and eight languages (Bengali, German, English, Indonesian, Korean, Portuguese, Russian, Mandarin). 
    }
    \label{fig:appedix-xgqa-figs}
\end{figure*}

Figure \ref{fig:appedix-xgqa-figs} shows the accuracy results for three evaluation setups (zero-shot, few shot learning and machine translation fine-tuning) on xGQA over the five different question types.
The language-specific sub-figures show the gap between xUNITER and TD-MML.
The largest differences between of them are the KOR, POR, RUS languages and for the \textit{compare}, \textit{logical}, and \textit{verify} question types.

\end{document}